\documentclass[journal,a4paper]{IEEEtran}
\usepackage{graphicx}
\usepackage{subfigure}
\usepackage{multirow}
\usepackage{array}
\usepackage{footnote}

\usepackage{algorithm}
\usepackage{algcompatible}
\usepackage{dsfont}
\usepackage{amsfonts,amsmath}
\usepackage{url,cite}
\usepackage{color}
\usepackage{bm}

\usepackage{booktabs}
\usepackage[normalem]{ulem}
\useunder{\uline}{\ul}{}
\usepackage[table,xcdraw]{xcolor}

\newcommand{\bfY}{\mathbf{Y}}
\newcommand{\bfy}{\mathbf{y}}

\newcommand{\bfA}{\mathbf{A}}
\newcommand{\bfa}{\mathbf{a}}

\newcommand{\bfv}{\mathbf{v}}
\newcommand{\bfV}{\mathbf{V}}

\newcommand{\bfw}{\mathbf{w}}
\newcommand{\bfW}{\mathbf{W}}

\newcommand{\bfz}{\mathbf{z}}
\newcommand{\bfZ}{\mathbf{Z}}

\newcommand{\bTheta}{\boldsymbol{\Theta}}

\newcommand{\MATabund}{\mathbf{A}}

\newcommand{\MATend}{\mathbf{S}}
\newcommand{\Vend}{\mathbf{s}}

\begin{document}

\title{AE-RED: A Hyperspectral Unmixing Framework Powered by Deep Autoencoder and Regularization by Denoising}

\author{Min~Zhao,~\IEEEmembership{Student Member,~IEEE,}
        Jie~Chen,~\IEEEmembership{Senior Member,~IEEE}
       and Nicolas Dobigeon,~\IEEEmembership{Senior Member,~IEEE}

       % <-this % stops a space

\thanks{M. Zhao and J. Chen are with School of Marine Science and Technology,
Northwestern Polytechnical University, Xi'an 710072, China (e-mail:
minzhao@mail.nwpu.edu.cn; dr.jie.chen@ieee.org).}
\thanks{N. Dobigeon is with University of Toulouse, IRIT/INP-ENSEEIHT,
CNRS, 2 rue Charles Camichel, BP 7122, 31071 Toulouse Cedex 7, France (e-mail: Nicolas.Dobigeon@enseeiht.fr).}
\thanks{Part of this work was supported by the Artificial Natural Intelligence Toulouse Institute (ANITI, ANR-19-PI3A-0004) and the IMAGIN project (ANR-21-CE29-0007).}}
%\markboth{IEEE xx,~Vol.~xx, No.~xx,~2020}%
%{Shell \MakeLowercase{\textit{et al.}}: Plug-and-Play Priors Framework for Hyperspectral Unmixing}
\maketitle
\begin{abstract}
Spectral unmixing has been extensively studied with a variety of methods and used in many applications.  Recently, data-driven techniques with deep learning methods have obtained great attention to spectral unmixing for its superior learning ability to automatically learn the structure information. In particular, autoencoder based architectures are elaborately designed to solve blind unmixing and model complex nonlinear mixtures. Nevertheless, these methods perform unmixing task as black-boxes and lack of interpretability. On the other hand, conventional unmixing methods carefully design the regularizer to add explicit information, in which algorithms such as plug-and-play (PnP) strategies utilize off-the-shelf denoisers to plug powerful priors. In this paper, we propose a generic unmixing framework to integrate the autoencoder network with regularization by denoising (RED), named AE-RED.
More specially, we decompose the unmixing optimized problem into two subproblems. The first one is solved using deep autoencoders to implicitly regularize the estimates and model the mixture mechanism. The second one leverages the denoiser to bring in the explicit information. 
In this way, both the characteristics of the deep autoencoder based unmixing methods and priors provided by denoisers are merged into our well-designed framework to enhance the unmixing performance. Experiment results on both synthetic and real data sets show the superiority of our proposed framework compared with state-of-the-art unmixing approaches.
\end{abstract}


\IEEEpeerreviewmaketitle
\begin{table}[h]
%\scriptsize
\centering
\renewcommand\arraystretch{1.15}
\caption{Notations.}\label{Tab_notation}
\begin{tabular}{ll}
$x$, $X$ & scalar \\
$\mathbf{x}$ & column vector \\
$\mathbf{X}$ & matrix \\
$B$ & number of spectral bands \\
$N$ & number of pixels \\
$R$ & number of endmembers \\
$\mathbf{y}_i\in\mathbb{R}^{B}$ & spectrum of the $i$th observed pixel \\
$\mathbf{Y}\in\mathbb{R}^{B\times N}$ & an observed hyperspectral image \\
$\mathbf{a}_i\in\mathbb{R}^{R}$ & abundance vector of the $i$th pixel \\
$\MATabund\in\mathbb{R}^{R\times N}$ & abundance matrix of all pixels \\
$\MATend\in\mathbb{R}^{B\times R}$ & endmember matrix with $R$ spectral signatures \\
$\boldsymbol{1}$ & all one vector or matrix \\
$\boldsymbol{0}$ & all zero vector or matrix\\
$\cdot\geq\cdot$ & elementwise inequality between vectors or matrices
\end{tabular}
\end{table}
\section{Introduction}
% problem formulation, general nonlinear model
\label{sec:intro} \IEEEPARstart{H}yperspectral imaging has been a widely explored imaging technique during recent years and is still receiving a growing attention in various applicative fields \cite{Dobigeon_IEEE_JSTARS_2014,yuen2010introduction}. Benefiting from the rich spectral information, hyperspectral images enable the analysis of fine materials in the observed scenes to tackle various challenging tasks such as target detection and classification~\cite{duan2021semisupervised,luo2020dimensionality}. However, due to the limitations of the imaging acquisition devices, there is an unsurmountable trade-off between the collected spectral and spatial information, which limits the spatial resolution of the hyperspectral sensors. As a consequence, a pixel observed by a hyperspectral sensor generally corresponds to a relatively large area and may encompass several materials, in particular when observing complex scenes. More precisely, the spectrum collected at a given spatial position of the scene is assumed to be a mixture of several elementary spectral signatures associated with the materials present in the observed pixel. This has led to research focused on hyperspectral unmixing (HU), which aims at decomposing the $i$th observed pixel spectrum $\bfy_i \in \mathbb{R}^B$ into a set of $R$ spectral signatures of so-called endmembers collected in the matrix $\MATend=\left[\mathbf{s}_1,\ldots,\mathbf{s}_R\right] \in \mathbb{R}^{B\times R}$ and their associated fractions or abundances $\bfa_i \in \mathbb{R}^{R}$~\cite{bioucas2013hyperspectral,dobigeon2013nonlinear,borsoi2021spectral}. 
%For a set of $N$ pixels sharing the same endmembers, assuming a linear mixing model (LMM) the observed spectral reflectance can be expressed as $\bfY=\mathcal{M}(\MATend,\bfa)+\bn$, where $\mathcal{M}(\cdot,\cdot)$ represents the mixing model between $\MATend$ and $\bfa$, and $\bn$ denotes the additive noise.
For the sake of physical interpretability, the abundances are subject to two constraints, namely abundance sum-to-one constraint (ASC), $\boldsymbol{1}_{R}^{\top}\bfa_i=1$, and abundance nonnegativity constraint (ANC), $\bfa_i\geq \boldsymbol{0}$. The endmembers are constrained to be nonnegative (ENC), $\MATend\geq \boldsymbol{0}$.
% traditional methods, deep methods

Many methods have been proposed in the literature to address the HU problem~\cite{dong2020spectral,zhao2021hyperspectralNMF,palsson2020convolutional,gao2021cycu,ozkan2018endnet}. Considering a set of $N$ observed pixels $\bfY=\left[\bfy_1,\ldots,\bfy_N\right] \in \mathbb{R}^{B\times N}$ sharing the same endmembers, HU can be formulated as an optimization problem, which aims at estimating the endmembers $\MATend$ and the abundances $\MATabund$ jointly, i.e.,  
% \begin{equation}
%\begin{split}\label{eq.general_model}
%            &\big\{\hat{\MATend}, \bTheta_{\MATend}^\star,\hat{\MATabund}, {\bTheta_{\mathbf{A}}^\star}, \hat{\mathcal{M}}\big\} =\\
%            &\mathop{\rm argmin}_{\MATend, \bTheta_{\MATend}, \MATabund, \bTheta_{\mathbf{A}}, \mathcal{M}\in\mathcal{F}} \quad \sum_{i=1}^N \mathcal{S}\big(\bfy_i,\hat{\bfY}_i\big) + \mathcal{R}\big(\MATend, \MATabund\big)\\
%           & \qquad {\rm with } \quad   \hat{\bfY}_i = {\mathcal{M}}\big(\MATend(\bTheta_{\MATend}), \bfa_i(\bTheta_{\mathbf{A}})\big) \\
%            & \qquad\;\; {\rm s.t. } \quad   \boldsymbol{1}_{R}^{\top}\MATabund=\boldsymbol{1}_{N}^{\top},~\MATabund\geq  \boldsymbol{0}, ~\text{and}~ \MATend\geq \boldsymbol{0}
%    \end{split}
%\end{equation}
 \begin{equation}
\begin{split}\label{eq.general_model}
            &\mathop{\rm min}_{\MATend,  \MATabund} \quad \sum_{i=1}^N \mathcal{D}\left[\bfy_i \big|\big| {\mathcal{M}}\big(\MATend, \bfa_i)\big)\right] + \mathcal{R}\big(\MATend, \MATabund\big)\\      
            & \qquad\;\; {\rm s.t. } \quad   \boldsymbol{1}_{R}^{\top}\MATabund=\boldsymbol{1}_{N}^{\top},~\MATabund\geq  \boldsymbol{0}, ~\text{and}~ \MATend\geq \boldsymbol{0}
    \end{split}
\end{equation}
where 
\begin{itemize}
\item $\mathcal{D}(\cdot,\cdot)$ stands for a discrepancy measure (e.g., divergence),
\item $\mathcal{M}(\cdot,\cdot)$ describes the inherent nonlinear mixture model which relates the endmembers and the abundances to the measurements, 
\item $\mathcal{R}(\cdot,\cdot)$ acts as a regularization term that encodes prior information regarding the endmembers $\MATend$ and the abundances $\MATabund$.
%\item $\bTheta_{\MATend}$ and $\bTheta_{\MATabund}$ denote the parameters to generate the endmembers $\MATend$ and abundances $\MATabund$, respectively.
%\item $\mathcal{F}$ is the function space where $\mathcal{M}$ lies in.
\end{itemize}
The regularization $\mathcal{R}(\cdot,\cdot)$ is often designed to be separable with respect to the abundances and endmembers,
\begin{equation}\label{eq.reg}
  \mathcal{R}(\MATend,\MATabund)=\mathcal{R}_{\rm e}(\MATend)+\mathcal{R}_{\rm a}(\MATabund),
\end{equation}
where the endmember and abundance prior information is encoded in $\mathcal{R}_{\rm e}$ and $\mathcal{R}_{\rm a}$, respectively. For instance, geometry-based penalizations, such as minimum volume~\cite{miao2007endmember}, are often chosen as endmember regularizers. Sparsity-based ~\cite{iordache2013collaborative}, low-rankness ~\cite{giampouras2016simultaneously} or spatial regularizers, such as total variation (TV)~\cite{iordache2012total}, are usually utilized to promote expected properties of the  abundances. This work  specifically focuses on the design of an abundance regularization. % and simply  define it as $\mathcal{R}(\MATabund)$.

As for the mixing process, typical methods rely on an explicit mathematical expression for $\mathcal{M}(\cdot,\cdot)$ to describe the mixture mechanism. For example, the linear mixing model (LMM) is by far the most used in the literature since it provides a generally admissible first-order approximation of the physical processes underlying the observations. LMM assumes that the measured spectrum is a linear combination of endmembers weighted by the abundances, which assumes that the incident light comes in and only reflects once on the ground before reaching the hyperspectral sensor. Besides, bilinear models consider second-order reflections, for instance in the case of multiple vegetation layers~\cite{halimi2011nonlinear,Dobigeon_IEEE_JSTARS_2014}. These explicit models are usually designed by describing the path of the light, along with its scattering and the interaction mechanisms among the materials. They are thus generally referred to as physics-based models. However, in some acquisition scenarios, they may fail to accurately account for real complex scenes. Data-driven methods have been thus proposed to implicitly learn the mixing mechanism from the observed data. Nevertheless, if not carefully designed a data-driven method may overlook the physical mixing process and require abundant training data~\cite{chen2023integration}.% To address these problems, integrating physics-based and data-driven models becomes a new trend.

\subsection{Motivation}
Numerous methods cope with the HU problem by carefully designing the data-fitting term and the regularizer~\cite{zhang2022sparse,peng2021low}. To reduce the computational complexity, most HU methods are based on the LMM. It may be not sufficient to account for spectral variability and endmember nonlinearity. On the other hand, designing a relevant regularizer is not always trivial and is generally driven by an empirical yet limited knowledge. For these reasons, research works have been devoted to the design of deep learning based HU approaches. Among them, autoencoders (AEs) become increasingly popular for unsupervised HU. The encoder is trained to compress the input into a lower dimensional latent representation, usually the abundances. The decoder is generally designed to mimic the  mixing process parametrized by the endmember signatures and to produce the hyperspectral image from the abundances defined in the latent space.
AE-based HU methods exhibit several advantages: \emph{i)} they can embed a physical-based mixing model into the structure of the decoder, \emph{ii)} they implicitly incorporates data-driven image priors and \emph{iii)} the unmixing procedure can benefit from powerful optimizers, such as Adam~\cite{kingma2014adam} and SGD~\cite{sutskever2013importance}. However, these deep architectures behave as black boxes and the results lack of interpretation. Motivated by these findings, this paper attempts to answer the following question: is it possible to design an unsupervised HU framework which combines the advantages of AE-based unmixing network while leveraging on explicit priors?

\subsection{Contributions}
This paper derives a novel HU framework which answers to this question affirmatively. More precisely, it introduces an AE-based unmixing strategy while incorporating an explicit regularization of the form of a RED. To solve the resulting optimization problem, an alternating direction method of multiplier (ADMM) is implemented with the great advantages of decomposing the initial problem into several simpler subproblems. One of these subproblems can be interpretated as a standard training task associated with an AE. Another is a standard denoising problem. The main advantages of the proposed frameworks are threefold:
\begin{itemize}
  \item This framework combines the deep AE with RED priors for unsupervised HU. By incorporating the benefits of AE with the regularization of denoising, the framework provides accurate unmixing results.
  \item The optimization procedure splits the unmixing task into two main subtasks. The first subtask involves training an AE to learn the mixing process and estimate a latent representation of the image as abundance maps. In the second subtask, a denoising step is applied to improve the estimation of the latent representation.
  \item The proposed framework is highly versatile and can accommodate various architectures for the encoder, and the decoder can be tailored to mimic any physics-based mixing model, such as the LMM, nonlinear mixing models, and mixing models with spectral variability.
\end{itemize}

This paper is organized as follows. Section~\ref{sec:related_work} provides a concise overview of related HU algorithms, with a particular focus on the design of regularizations and AE-based unmixing methods. Section~\ref{sec:background} describes some technical ingredients necessary to build the proposed framework. In Section~\ref{sec:proposed}, the proposed generic framework is derived, and details about particular instances of this framework are given. Section~\ref{sec:experiment} reports the results obtained from extensive experiments conducted on synthetic and real datasets to demonstrate the superiority of the proposed framework. Finally, Section~\ref{sec:conclusion} concludes the paper.

\section{Related works}
\label{sec:related_work}
This section provides brief overviews on two aspects related to this work, namely regularization design in HU and AE-based unmixing.

\subsection{Regularization design}

Efficient algorithms for HU often require effective regularizations that incorporate prior knowledge about the images and constrain the solution space. Traditional methods exploit the spatial consistency of the image, and sparsity-based regularizers have also been extensively used on the abundances since the number of endmembers is typically much smaller than the size of the spectral library.

In~\cite{iordache2012total}, a TV regularizer is applied to the abundance to promote similarity between adjacent pixels, and an $\ell_1$-norm is used for sparse unmixing. Since the $\ell_1$-norm is inconsistent with the abundance sum-to-one constraint, $\ell_p$-norms with $0<p<1$ have been studied to obtain sparse estimates~\cite{sigurdsson2014hyperspectral}. In~\cite{zhong2013non}, a non-local sparse unmixing method is proposed to exploit similar patterns and structures in the abundance image. A weighted average is applied to all pixels to exploit non-local spatial information. A weighted average is applied to all pixels to exploit non-local spatial information. Spatial group sparsity regularizers have also been proposed to incorporate spatial priors and sparse structures. The authors of~\cite{wang2017spatial} introduce a spatial group sparsity regularizer generated using image segmentation methods such as SLIC. In~\cite{lagrange2020matrix}, a cofactorization model is used to jointly exploit spectral and spatial information, while the work of~\cite{tong2020adaptive} introduces an adaptive graph to automatically determine the best neighbor points of pixels and assign corresponding weights.  However, these methods require handcrafted regularizers, which can be time-consuming when non-standard regularizers are applied to large images.

More recently, the idea of PnP has been introduced to exploit the intrinsic properties of hyperspectral images. These methods use generic denoisers that act as explicit regularizers. In~\cite{zhao2021plug}, an HU method based on an ADMM algorithm is introduced that can handle explicit regularizations. By selecting different pattern switch matrices, the denoising operator can be used to penalize the reconstructed hyperspectral image or estimated abundances. The work of~\cite{wang2020hyperspectral} proposes a nonlinear unmixing method with prior information provided by denoisers. However, the denoisers used in these methods are traditional denoising methods or deep denoisers trained on grayscale or RGB images, which may not be optimal for hyperspectral images.

\subsection{Deep AE-based unmixing methods}
Elegant neural network structures have been proposed to formulate the HU task as a simple training process. Early works used fully connected layers to design the model, such as~\cite{ozkan2018endnet} and~\cite{qu2018udas}. However, these networks process the pixels independently and ignore the spatial correlation intrinsic to the image.  To overcome this limitation, some AE-based methods include spatial regularizations, such as total variation (TV), in the loss function~\cite{zhao2021lstm}. Recently, convolutional neural networks (CNNs) have been used to perform HU and have shown promising performance. CNNs convolve the input data with filter kernels to capture spatial information~\cite{palsson2020convolutional,zhao2021hyperspectral}.  Recurrent neural networks (RNNs), which have memory cells, implement a sequential process with hidden states that depend on the previous states~\cite{zhao2021lstm}. Hyperspectral images are often corrupted by noise or outliers, which can dramatically decrease the unmixing performance. To address this issue, denoising-oriented architectures have been proposed~\cite{qu2018udas}.  Some works have also proposed variants of encoders. In~\cite{hua2021dual}, a dual-branch AE network is designed to leverage multiscale spatial contextual information.

Most AE-based HU methods use a fully connected linear layer in the decoder part to mimic the linear mixing process. However, considering the physical interactions between multiple materials and the superior ability of deep networks to model nonlinear relationships, some works~\cite{NAE2019,zhao2021hyperspectral,zhao2021lstm,shahid2021unsupervised} have focused on the design of structured decoders to ensure the interpretability of the nonlinear model inherent in the mixing process.  The work of~\cite{NAE2019} introduces a nonlinear decoder. Recycling an LMM-based AE architecture, the decoder contains two parts: one linear and the other nonlinear. The linear part is considered a rough approximation of the mixture and is then fed into two fully connected layers with a nonlinear activation function to learn the nonlinear mechanism. However, this post-nonlinear model-based decoder may not be sufficient to represent complex nonlinear cases.  Some works~\cite{zhao2021hyperspectral,zhao2021lstm,shahid2021unsupervised} reexamine the nonlinear fluctuation part of the decoder. For example, the method in~\cite{shahid2021unsupervised} designs a special layer to capture the second-order interaction, similar to the Fan or bilinear models. Moreover, spectral variability can also be addressed by using deep generative decoders~\cite{borsoi2019deep,shi2021probabilistic}.

Recently, deep unfolding techniques have been used to unroll a model and its related iterative algorithm into deep networks. This approach can include physical interpretability into the design of network layers, and such model-inspired networks are also used in the design of unmixing methods.  In~\cite{qian2020spectral}, an iterative shrinkage-thresholding algorithm (ISTA)-inspired network layer is applied to build an AE-based unmixing architecture. The work of~\cite{xiong2021snmf} unrolls a sparse non-negative matrix factorization (NMF)-based algorithm with an $\ell_p$-norm regularizer to integrate prior knowledge into the unmixing network. An ADMM solver with a sparse regularizer is also unrolled to build an AE-like unmixing architecture.  However, these methods do not utilize spatial consistency information in the design of the network, which may limit their unmixing performance.

\begin{figure*}[!t]
  \centering
  \includegraphics[width=18cm]{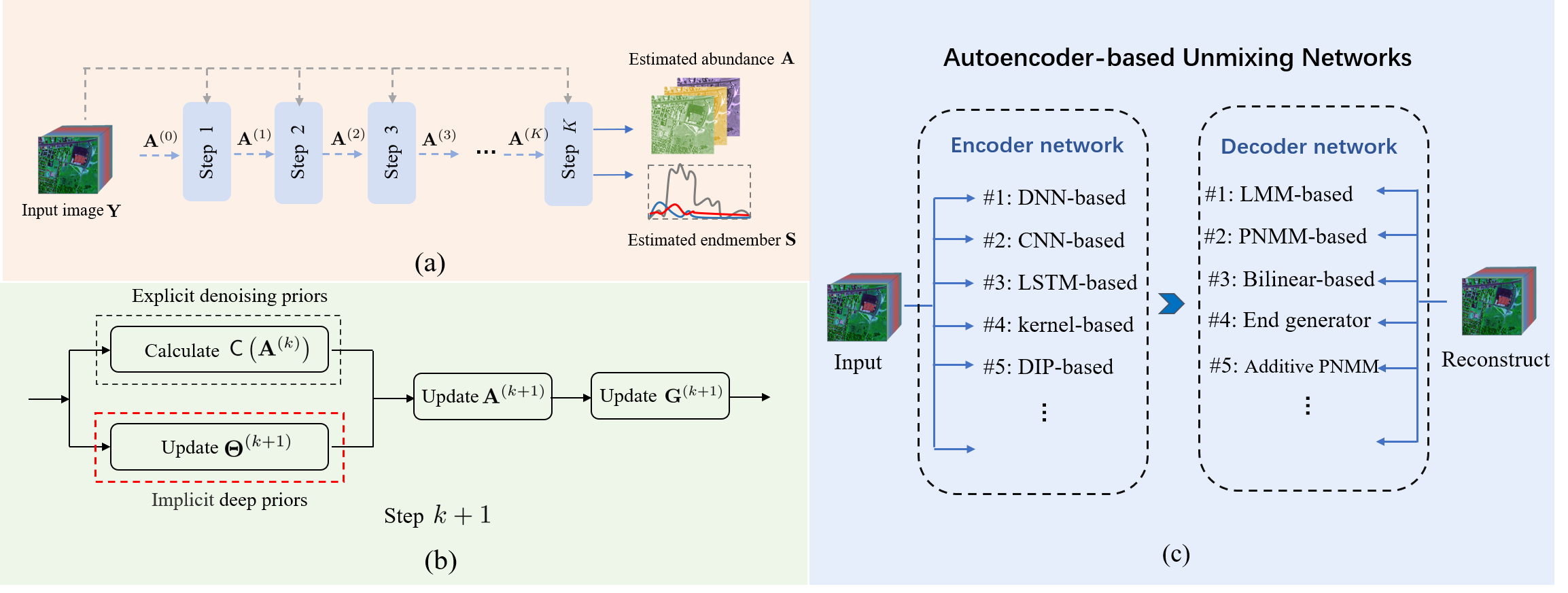}\\
  \caption{Framework of the proposed AE-RED. (a) The scheme of the proposed framework. (b) Flowchart of the $(k+1)$th ADMM step: the denoising operator is applied in parallel to the update of $\bTheta$ to speed up calculations. (c) An overview and some instances of AE-based unmixing networks, where the encoder embeds deep priors for abundance estimation, and the decoder can model the mixture mechanism and extract the endmembers. The choice of the encoder and decoder is let to the end-user. }\label{fig.framework}
\end{figure*}

\section{Background} \label{sec:background}
\subsection{Autoencoder-based unmixing}
As highlighted in the previous section, AEs have demonstrated to be a powerful tool to conduct unsupervised unmixing. An AE typically consists of an encoder and a decoder. The encoder, represented by $\mathsf{E}_{\bTheta_{\sf E}}(\cdot)$, aims at learning a nonlinear mapping from input data, denoted as $\bfw_i$, to their corresponding latent representations, denoted as $\bfv_i$. This can be expressed as follows:
\begin{equation}\label{eq.encoder}
  \bfv_i= \mathsf{E}_{\bTheta_{\sf E}}(\bfw_i),
\end{equation}
where $\bTheta_{\sf E}$ gather all parameters of the encoder. The input $\bfW=\left[\bfw_1,\ldots,\bfw_N\right]$ depends on the architecture chosen for the encoder network. For instance, when dealing with the specific task of HU, the input can be chosen as the image pixels $\bfY=\left[\bfy_1,\ldots,\bfy_N\right]$ or as noise realizations $\bfZ=\left[\bfz_1,\ldots,\bfz_N\right]$ with $\bfz_i \sim \mathcal{N}(\boldsymbol{0},\mathbf{I})$. The decoder, denoted by $\mathsf{D}_{\bTheta_{\sf D}}(\cdot)$, is responsible for reconstructing the data, or at least an approximation $\hat{\bfy}_i$, from the latent feature $\bfv_i$ provided by the encoder. This can be expressed as follows:
\begin{equation}\label{eq.decoder}
  \hat{\bfy}_i= \mathsf{D}_{\bTheta_{\sf D}}(\bfv_i),
\end{equation}
where $\bTheta_{\sf D}$ parameterizes the decoder. Under this paradigm, adjusting the encoder and decoder parameters $\bTheta_{\sf E}$ and $\bTheta_{\sf D}$ is generally achieved by minimizing the empirical expectation of a discrepancy measure between the input data $\bfy_1,\ldots,\bfy_N$ and their corresponding approximation $\hat{\bfy}_1,\ldots,\hat{\bfy}_N$, i.e.,
\begin{equation}\label{eq.loss_ae}
  \mathcal{L}(\bTheta_{\sf E},\bTheta_{\sf D})=\frac{1}{N}\sum_{i=1}^{N}\mathcal{D}\left[\mathbf{y}_i||
  \hat{\bfy}_i\right]
\end{equation}
with $\hat{\bfy}_i= \mathsf{D}_{\bTheta_{\sf D}}\left(\mathsf{E}_{\bTheta_{\sf E}}\left(\bfw_i\right)\right)$. This reconstruction loss function can be complemented with additional terms to account for any desired property regarding the network parameters and the latent representation. %
%The objective function of autoencoder is usually designed to enforce the network to reconstruct the input pixel by minimizing the distance between original pixel $\bfY$ and its corresponding reconstruction $\boldsymbol{\widehat{y}}$. Using the mean-squared error as an example, its mathematical form can be given by,
%\begin{equation}\label{eq.loss_ae}
%  \mathcal{L}_\text{MSE}=\frac{1}{N}\sum_{i=1}^{N}\|\mathbf{y}_i-
%  \boldsymbol{\widehat{y}}_i\|_2^2.
%\end{equation}

Drawing a straightforward analogy with the problem \eqref{eq.general_model}, AE-based unmixing frameworks generally assume that the latent variable $\bfV=\left[\bfv_1,\ldots,\bfv_N\right]$ can be considered as an estimate of the abundance matrix $\bfA$. The architecture of the encoder should be chosen to be able to extract key spatial features from the input data. Several choices are possible and will be discussed as archetypal examples later in Section \ref{sec:proposed-B}. The decoder can then be designed to mimic the mixing process $\mathcal{M}(\cdot,\cdot)$ in \eqref{eq.general_model}. The endmember signatures to be recovered are part of the decoder parameters, i.e., $\boldsymbol{\Theta}_{\sf D} = \left\{\tilde{\boldsymbol{\Theta}}_{\sf D},\MATend\right\}$ where $\tilde{\boldsymbol{\Theta}}_{\sf D}$ are intrinsic network parameters. For instance, when the decoder is designed according to a physics-based nonlinear mixing model prescribed beforehand, $\tilde{\boldsymbol{\Theta}}_{\sf D}$ gathers the nonlinearity parameters. In the simplistic assumption of the LMM, the decoder does not depend on any additional intrinsic parameters and  $\boldsymbol{\Theta}_{\sf D} = \MATend$.

%  The parameters of a special designed layer of decoder can be considered as endmembers. This framework can implicitly incorporate the inference relation from the observed spectra to abundances and endmembers with the priors embedded in the network parameters. With the introduction of layers considered spatial information, e.g. CNN, the encoder acts as a regularizer relating the estimation of abundances.
% The decoder mimics the mixture model and can exploit data-driven strategy to model complex mixture scenes.

\subsection{Regularization by denoising priors}
Various regularizers have been considered to design the term $\mathcal{R}_{\mathrm{a}}(\cdot)$. Among them, PnP is a flexible and generic framework that naturally emerges when resorting to splitting-based optimization procedures. This framework replaces the proximal operator associated with $\mathcal{R}_{\mathrm{a}}(\cdot)$ by off-the-shelf and highly engineered denoiser. This strategy has been effectively used when tackling many imaging inverse problems, such as image denoising, super-resolution and inpainting \cite{lai2022deep,dian2020regularizing}. Recently, an advanced version of PnP, regularization by denoising (RED)~\cite{romano2017little} has demonstrated superior performance. It can be expressed as
\begin{equation}\label{eq.red}
  \mathcal{R}_{\mathrm{a}}(\MATabund)= \frac{1}{2}\MATabund^{\top}\left(\MATabund-\mathsf{C}\left(\MATabund\right)\right),
\end{equation}
where $\mathsf{C}(\cdot)$ is a denoiser. This regularizer is proportional to the inner-product between the abundance and its post-denoising residual and exhibits many appealing characteristics. First, it is a convex function. Second, under some mild assumptions and reasonable conditions on $\mathsf{C}(\cdot)$, its derivative with respect to $\MATabund$ is simple and given as the denoising residual, i.e., $\nabla\mathcal{R}(\MATabund)=\MATabund-\mathsf{C}\left(\MATabund\right)$~\cite{romano2017little}. This work aims at devising a generic AE-based HU framework that can incorporate the RED regularizer.

\section{Proposed method}
\label{sec:proposed}
%framework

\subsection{Generic framework}
% linear model, additive model, post nonlinear model
The generic unmixing framework proposed in this paper, referred to as AE-RED hereafter, formulates the HU problem as the training of an AE while leveraging  the RED paradigm. Adopting a conventional Euclidean divergence for $\mathcal{D}(\cdot,\cdot)$, the HU problem
\eqref{eq.general_model} is now specified as
\begin{equation}
\begin{aligned}\label{eq_mdoel_2D_RED}
  \min_{\bTheta}&
  \left\|\mathbf{Y}-\mathsf{D}_{\bTheta_{\sf D}}\left(
  \mathsf{E}_{\bTheta_{\sf E}}(\mathbf{W})\right)\right\|_{\text{F}}^2\\
  &+\lambda \mathsf{E}_{\bTheta_{\sf E}}(\mathbf{W})^\top\left(\mathsf{E}_{\bTheta_{\sf E}}(\mathbf{W})-\mathsf{C}\left(\mathsf{E}_{\bTheta_{\sf E}}(\mathbf{W})\right)\right)\\
  \text{s.t.}\  &\boldsymbol{1}^{\top}_{R}\mathsf{E}_{\bTheta_{\sf E}}(\mathbf{W})=\boldsymbol{1}_{N}^{\top},~\mathsf{E}_{\bTheta_{\sf E}}(\mathbf{W})\geq \boldsymbol{0}~\text{and}~\MATend \geq \boldsymbol{0}
\end{aligned}
\end{equation}
with $\bTheta = \left\{\bTheta_{\sf E}, \bTheta_{\sf D}\right\}$. As stated in the previous section, the endmembers are part of the set of decoder parameters, i.e., $\bTheta_{\sf D}=\left\{\tilde{\bTheta}_{\sf D},\MATend\right\}$ and the latent representation directly provides abundance estimates, i.e., $\mathbf{A}=\mathsf{E}_{\bTheta_{\sf E}}(\mathbf{W})$. This formulation of the unmixing task leverages on a combination of the AE modeling and RED, providing several benefits. First, the AE is effective in handling the mixture mechanism and learning underlying information. Second, RED provides a flexible and efficient way to model data priors.

Solving the minimization problem \eqref{eq_mdoel_2D_RED} with deep learning-flavored black-box optimizers is challenging if not infeasible, in particular because back-propagating $\bTheta_{\sf E}$ would require differentiating the denoising function $\mathsf{C}$. For most denoisers, this differentiation is not straightforward and may need a huge amount of computations. However, it is worth noting that one of the great advantages of RED is that its  derivative can be directly calculated. To benefit from this property, one simple strategy consists in reintroducing the abundance matrix $\MATabund$ explicitly as an auxiliary variable and then reformulating \eqref{eq_mdoel_2D_RED} as a constrained problem
\begin{equation}
\begin{aligned}\label{eq_mdoel_2D_RED_2}
  \min_{\bTheta, \MATabund}
  &\left\|\mathbf{Y}-\mathsf{D}_{\bTheta_{\sf D}}\left(
  \mathsf{E}_{\bTheta_{\sf E}}(\mathbf{W})\right)\right\|_{\text{F}}^2+\lambda{\MATabund}^\top\left(\MATabund-\mathsf{C}\left(\MATabund\right)\right)\\
    \text{s.t.}\  &\boldsymbol{1}^{\top}_{R}\mathsf{E}_{\bTheta_{\sf E}}(\mathbf{W})=\boldsymbol{1}_{N}^{\top},~\mathsf{E}_{\bTheta_{\sf E}}(\mathbf{W})\geq \boldsymbol{0},~\MATend \geq \boldsymbol{0}\\
    \text{and}\  & \MATabund =\mathsf{E}_{\bTheta_{\sf E}}(\mathbf{W}). 
%  \text{s.t.}\ &,~ \boldsymbol{1}^{\top}_{R}\mathsf{E}_{\bTheta_{\sf E}}(\mathbf{Z})=\boldsymbol{1}_{N}^{\top},\\
%  &\mathsf{E}_{\bTheta_{\sf E}}(\mathbf{Z})\geq \boldsymbol{0}~\text{and}~\MATend \geq \boldsymbol{0}.
\end{aligned}
\end{equation}
% The corresponding augmented Lagrangian function is:
% \begin{equation}
% \begin{aligned}\label{eq_mdoel_2D_RED_L}
%   &\mathcal{L}\left(\bTheta_{\sf E}, \bTheta_{\sf D}, \MATabund, \mathbf{G} \right)=
%   \|\mathbf{Y}-\mathsf{D}_{\bTheta_{\sf D}}\left(
%   \mathsf{E}_{\bTheta_{\sf E}}(\mathbf{W})\right)\|_{\text{F}}^2\\
%   +\lambda{\MATabund}^\top&\left(\MATabund-\mathsf{C}\left(\MATabund\right)\right)+\mu\|\MATabund-\mathsf{E}_{\bTheta_{\sf E}}(\mathbf{W})-\mathbf{G}\|_{\text{F}}^{2},\\
%   ~~~\text{s.t.}~&\boldsymbol{1}^{\top}_{R}\mathsf{E}_{\bTheta_{\sf E}}(\mathbf{W})=\boldsymbol{1}_{N}^{\top},~\mathsf{E}_{\bTheta_{\sf E}}(\mathbf{W})\geq \boldsymbol{0}~\text{and}~\MATend \geq \boldsymbol{0},
% \end{aligned}
% \end{equation}
To solve  \eqref{eq_mdoel_2D_RED_2}, a common yet efficient strategy boils down to split the initial problems into several simpler subproblems following an ADMM. The main steps of the resulting algorithmic scheme write
%\begin{equation}
%\begin{aligned}\label{eq_mdoel_2D_RED_T}
%&\bTheta^{(k)}=\arg\min_{\bTheta}\|\bfY-\mathsf{D}_{\bTheta_{\sf D}}\left(
%  \mathsf{E}_{\bTheta_{\sf E}}(\mathbf{W})\right)\|_{\text{F}}^2\\
%  &~~~~~~~~~~~~~+\mu\|\MATabund^{(k-1)}-\mathsf{E}_{\bTheta_{\sf E}}(\mathbf{W})-\mathbf{G}^{(k-1)}\|_{\text{F}}^{2},\\
%  &\text{s.t.}~\boldsymbol{1}^{\top}_{R}\mathsf{E}_{\bTheta_{\sf E}}(\mathbf{W})=\boldsymbol{1}_{N}^{\top},~ \mathsf{E}_{\bTheta_{\sf E}}(\mathbf{W})\geq \boldsymbol{0}~\text{and}~\MATend\geq %\boldsymbol{0}\\
%  \end{aligned}
%\end{equation}
\begin{align}
\bTheta^{(k+1)}&=\arg\min_{\bTheta}\|\bfY-\mathsf{D}_{\bTheta_{\sf D}}\left(\mathsf{E}_{\bTheta_{\sf E}}(\mathbf{W})\right)\|_{\text{F}}^2  \label{eq_mdoel_2D_RED_T} \\
               &+\mu\|\MATabund^{(k)}-\mathsf{E}_{\bTheta_{\sf E}}(\mathbf{W})-\mathbf{G}^{(k)}\|_{\text{F}}^{2} \nonumber \\
  \text{s.t.}  &\ \boldsymbol{1}^{\top}_{R}\mathsf{E}_{\bTheta_{\sf E}}(\mathbf{W})=\boldsymbol{1}_{N}^{\top},~ 
                    \mathsf{E}_{\bTheta_{\sf E}}(\mathbf{W})\geq \boldsymbol{0}~\text{and}~\MATend\geq \boldsymbol{0} \nonumber\\
\mathbf{A}^{(k+1)}&= \arg\min_{\MATabund}\lambda \MATabund^\top\left(\MATabund-\mathsf{C}\left(\MATabund\right)\right)  \label{eq_mdoel_2D_RED_A} \\
               &+\mu\|\MATabund-\mathsf{E}_{\bTheta_{\sf E}}^{(k+1)}(\mathbf{W})-\mathbf{G}^{(k)}\|_{\text{F}}^{2} \nonumber \\
\mathbf{G}^{(k+1)}&=\mathbf{G}-\MATabund^{(k+1)}+\mathsf{E}_{\bTheta_{\sf E}}^{(k+1)}(\mathbf{W}) \label{eq_mdoel_2D_RED_g} 
\end{align}
where  $\mu$ is the penalty parameter and $\mathbf{G}$ is the dual variable.
%The endmembers and abundances can be estimated by iteratively updating $\bTheta_{\sf E}$, $\bTheta_{\sf D}$, $\MATabund$ and $\mathbf{G}$. Note that $\bTheta_{\sf E}$ and $\bTheta_{\sf D}$ are related to the parameters of an autoencoder, so we can update them simultaneously in the training of network. It is actually optimized separately during gradient derivation.
The framework of the proposed AE-RED is summarized in Fig.~\ref{fig.framework}. It embeds a data-driven autoencoder with a model-free RED. The algorithmic scheme is shown to be a convenient way to fuse the respective advantages of these two approaches. Note that, since the AE-based formulation is nonlinear, providing convergence guarantees about the resulting optimization scheme is not trivial. However, the experimental results reported in Section~\ref{sec:experiment} show that the proposed method is able to provide consistent performance. Finally, without loss of generality, detailed technical implementations of the first two steps \eqref{eq_mdoel_2D_RED_T} and \eqref{eq_mdoel_2D_RED_A} are discussed in the following paragraphs for specific architectures of the autoencoder.

\begin{figure*}[!t]
  \centering
  \includegraphics[width=18cm]{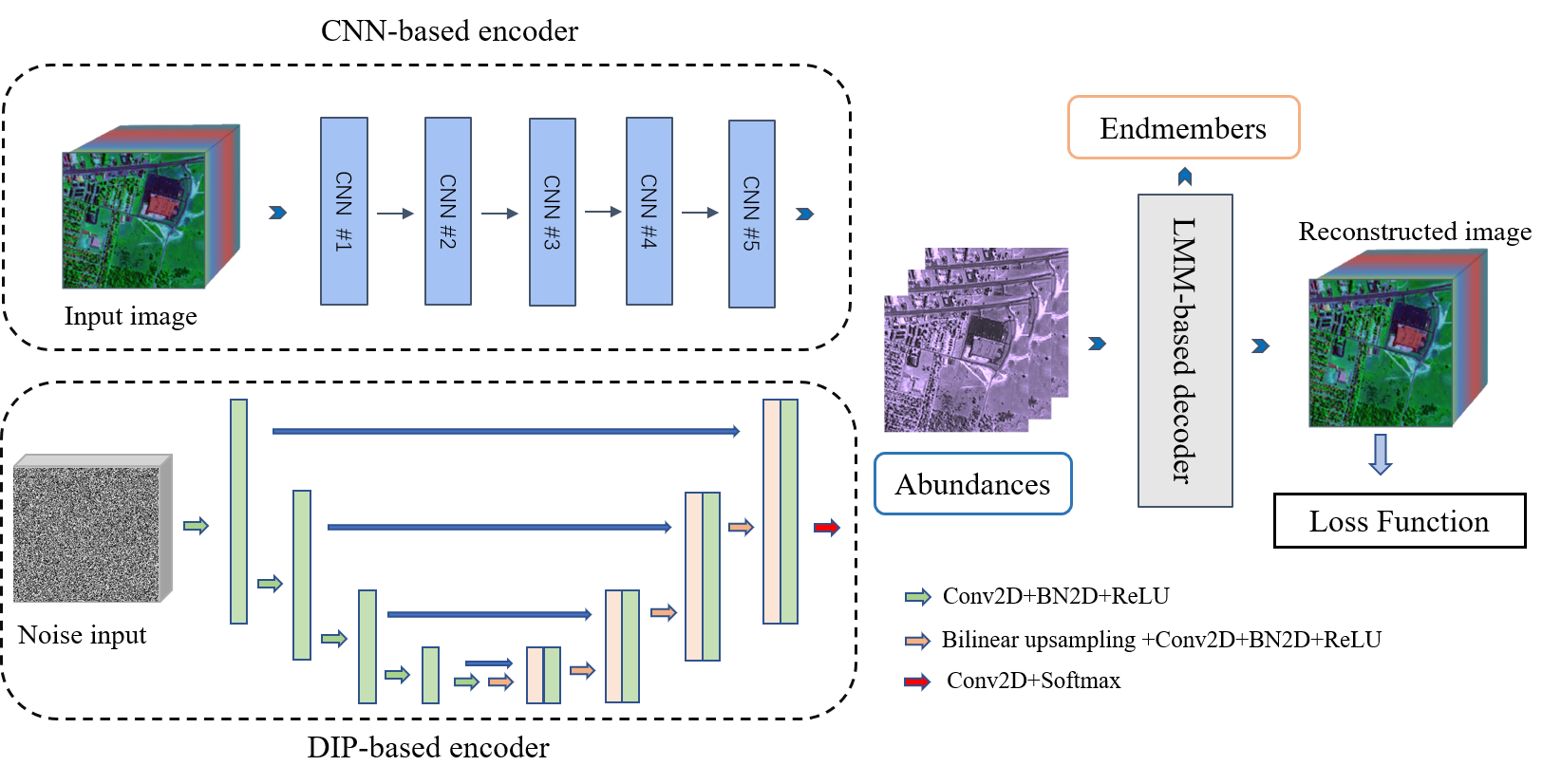}\\
  \caption{The architectures of CNN-based and DIP-based networks used as particular instances of the proposed method. }\label{fig.network}
\end{figure*}

\subsection{Updating $\boldsymbol{\Theta}$}
\label{sec:proposed-B}
At each iteration, the set of parameters $\boldsymbol{\Theta}$ of the autoencoder is updated through rule \eqref{eq_mdoel_2D_RED_T}. This can be achieved by training the network with the function in \eqref{eq_mdoel_2D_RED_T} as the objective function. The first term measures the data fit while the second acts as a regularization to enforce the representation  $\mathsf{E}_{\bTheta_{\sf E}}(\mathbf{W})$ in the latent space to be close to a corrected version $\MATabund-\mathbf{G}$ of the abundance. %This proximity regularization is to fuse the explicit priors with deep features, which can make the network minimization stable and improve the unmixing performance. 
Regarding the ASC, ANC and ENC constraints, they can be ensured by an appropriate design of the network. In practice, Adam is used to train the autoencoder.

Various autoencoder architectures can be envisioned and the encoder and the decoder can be chosen by the end-user with respect to the targeted applicative context. The encoder $\mathsf{E}_{\bTheta_{\sf E}}(\cdot)$ aims at extracting relevant features to be incorporated into the estimated abundances. A popular choice is a CNN-based architecture where the input is the observed image. Another promising approach consists in leveraging on a deep image prior (DIP) with a noise input. These two particular choices are discussed later in this section. Regarding the decoder $\mathsf{D}_{\bTheta_{\sf D}}(\cdot)$, it generally mimics the mixing process and the endmembers usually define the weights of one specially designed linear layer. Again, the proposed AE-RED framework is sufficiently flexible to host various architectures and to handle various spectral mixing models. A popular strategy is to design the decoder such that it combines physics-based and data-driven strategies to account for complex nonlinearities or spectral variabilities. For instance, additive nonlinear and post-nonlinear models have been extensively investigated \cite{zhao2021hyperspectral,zhao2021lstm,shahid2021unsupervised}  as well as spectral variability-aware endmember generators \cite{borsoi2019deep,shi2021probabilistic}.

Some archetypal examples of possible elements composing the architecture of the AE are (non-exhaustively) listed in Fig.~\ref{fig.framework}(c). In the sequel of this paper, for illustration purpose but without loss of generality, two particular architectures are discussed and then instantiated, as shown in Fig.~\ref{fig.network}. Both consider an LMM-based decoder composed of a convolutional layer with a filter size of $1\times 1\times B$ to mimic the LMM. The adjusted decoder weights are finally extracted to estimate the endmember spectral signature. For this particular instance of the decoder, the optimization problem  \eqref{eq_mdoel_2D_RED_T} can be rewritten as
\begin{align}\label{eq_mdoel_2D_RED_T_2}
  \left\{\bTheta_{\sf E}, 
  \mathbf{S}\right\}&\in\arg\min_{\bTheta_{\sf E}, 
  \mathbf{S}}\|\bfY-\mathbf{S}
  \mathsf{E}_{\bTheta_{\sf E}}(\mathbf{W})\|_{\text{F}}^2\\
  &+\mu\|\MATabund-\mathsf{E}_{\bTheta_{\sf E}}(\mathbf{W})-\mathbf{G}\|_{\text{F}}^{2} \nonumber\\
  \text{s.t.}& \ \boldsymbol{1}^{\top}_{R} \mathsf{E}_{\bTheta_{\sf E}}(\mathbf{W})=\boldsymbol{1}_{N}^{\top},~ \mathsf{E}_{\bTheta_{\sf E}}(\mathbf{W})\geq \boldsymbol{0}~\text{and}~\MATend\geq \boldsymbol{0}. \nonumber
\end{align}
The two examples of AE considered in this paper differ by the architecture of the encoder. The first network is composed of a CNN-based encoder while the second is a DIP. These two choices are discussed below.

\subsubsection{CNN-based encoder}
The architecture of the CNN-based encoder is shown in Fig.~\ref{fig.network}. The whole image $\mathbf{Y}$ is used here as the input to extract the structure information from the hyperspectral image. Another choice would consist in considering over-lapping patches as the input. The encoder is composed of 5 blocks. The first two blocks implement $3\times 3$ convolution filters to learn the spatial consistency information. The next two blocks apply $1\times 1$ convolution operators (i.e., fully connected layers) to model the spectral priors. Moreover, to satisfy the ANC and ASC, the conventional LeakyReLU activation function of the last block is replaced by a softmax function. The output dimensions of each block are narrowly diminished to compress the input pixels into the abundance domain. Considering the optimization function defined in~\eqref{eq_mdoel_2D_RED_T_2}, the objective function to train this model is expressed as
\begin{equation}\label{eq_loss}
  \mathcal{L}_{\text{AE}}(\boldsymbol{\Theta})=\|\mathbf{Y}-\MATend
  \mathsf{E}_{\bTheta_{\sf E}}(\bfY)\|_{\text{F}}^2 +\mu\|\MATabund-\mathsf{E}_{\bTheta_{\sf E}}(\bfY)-\mathbf{G}\|_{\text{F}}^{2}.
\end{equation}
The resulting unmixing method will be denoted as AE-RED-C in the sequel.

\subsubsection{Deep image prior-based encoder}
Another architecture considered in this paper exploits the DIP strategy  to implicitly learn the priors of hyperspectral image. Unlike conventional AE-based unmixing methods which use spectral signatures as input for training, this network applies a Gaussian noise image $\mathbf{Z}$ of size of the abundance matrix $\MATabund$ as input to generate the hyperspectral image. The encoder can be a U-net like architecture to extract the features from different levels. In this work the encoder has been designed with an encoder-decoder structure for abundance estimation. The inner encoder is composed of 4 down-sampling to compress the features. Each down-sampling block consists of three layers, namely a convolution layer with a filter of size $3\times 3$, a batch normalization layer, and a ReLU nonlinear activation layer. The inner decoder is composed of 5 up-sampling blocks. Each of the first 4 blocks has 4 layers: a bilinear upsampling layer, a convolution layer, a batch normalization layer and a ReLU nonlinear activation layer. The last block has two layers, namely a convolution layer and a softmax nonlinear activation layer to generate the estimated abundances while satisfying the ANC and ASC. Skip connections relate the encoder and decoder which are used to fuse the low-level and high-level features and to obtain  multiscale information. The objective function to train this deep model is also defined as \eqref{eq_loss} where $\mathsf{E}_{\bTheta_{\sf E}}(\bfY)$ is replaced by $\mathsf{E}_{\bTheta_{\sf E}}(\bfZ)$.
The proposed method with this architecture is denoted as AE-RED-U.

\subsection{Updating $\MATabund$}
The abundance matrix $\MATabund$ is updated by solving \eqref{eq_mdoel_2D_RED_A}.
This problem is a standard RED objective function and can be interpreted as a denoising of $\mathsf{E}_{\bTheta_{\sf E}}(\mathbf{W})+\mathbf{G}$. The seminal paper \cite{romano2017little} discusses two algorithmic schemes to solve this problem, namely fixed-point and gradient-descent strategies. In this work we derive a fixed-point algorithm by setting the gradient of the objective function to $\boldsymbol{0}$,
\begin{equation}
\begin{aligned}\label{eq_mdoel_2D_RED_Ag}
  \lambda \left(\MATabund-\mathsf{C}\left(\MATabund\right)\right)+\mu\left(\MATabund-\mathsf{E}_{\bTheta_{\sf E}}(\mathbf{W})-\mathbf{G}\right)=\boldsymbol{0}.
\end{aligned}
\end{equation}
Then, at the $(k+1)$th iteration of the ADMM, the $j$th inner iteration of the fixed-point algorithm can be summarized as
\begin{multline}
\label{eq_mdoel_2D_RED_Ag_f}
  \MATabund^{(k+1,j+1)}\\
  =\frac{1}{\lambda+\mu}\left[\lambda \mathsf{C}\left(\MATabund^{(k+1,j)}\right)
  +\mu\left(\mathsf{E}^{(k+1)}_{\bTheta_{\sf E}}(\mathbf{W})+\mathbf{G}^{(k)}\right)\right].
\end{multline}
%The update of $\MATabund$ is conducted $J$ times to provided a needed solution. The denoiser operator $\mathsf{C}(\cdot)$ can be conducted in parallel to the update of $\boldsymbol{\Theta}$ and speed up the running time of our proposed method. 
For illustration, we consider two particular denoisers $\mathsf{C}\left(\cdot\right)$, namely nonlocal means (NLM) \cite{buades2011non} and block-matching and 4-D filtering (BM4D) \cite{maggioni2012nonlocal}. NLM is a 2D denoiser and should be applied on each spectral bands indendently while and BM4D is a 3D-cube based denoiser. Depending on the architecture chosen for the encoder (see Section \ref{sec:proposed-B}), the corresponding instances of the proposed framework are named as AE-RED-CNLM, AE-RED-CBM4D, AE-RED-UNLM and AE-RED-UBM4D, respectively.
%denoiser

%\subsection{Updating $\mathbf{G}$}
%The Lagrange multiplier $\mathbf{G}$ is calculated by \eqref{eq_mdoel_2D_RED_g}.

%Our AEPnP method is a combination of autoencoder network and PnP regularization, where the autoencoder has superior capability in handling the mixture mechanism and learning the underlying information, and the PnP provides a flexible and effective way to model explicit prior. 
%Our method can be seen as an ensemble framework to integrate the data-driven methods and prior model algorithms, where the ADMM is applied to fuse the characteristics of these two terms. 
%Note that as the autoencoder based unmixing system is nonlinear, it is nontrival to provide the convergence guarantees of our optimization framework. But from the experiment results in Section~\ref{sec:experiment}, we can observe that our AEPnP method is convergent and can get good unmixing results.

\newcommand{\algocomment}[1]{ \STATEx {\color[rgb]{0.5,0.8,0.5}{\% \textit{#1}}}}

\begin{algorithm}[!t]
	\renewcommand{\algorithmicrequire}{\textbf{Input:}}
	\renewcommand{\algorithmicensure}{\textbf{Output:}}
	\caption{The proposed unmixing framework AE-RED}
	\label{alg:1}
	\begin{algorithmic}[1]
	\REQUIRE Hyperspectal image $\mathbf{Y}$; Regularization parameter $\lambda$; ADMM coefficient $\mu$; Denoiser $\mathsf{C}(\cdot)$; Outer and inner iteration numbers $K$ and $J$; Training parameters (learning rate, epochs, batch size).
%     \STATE \textbf{Initialization:}\\
% \begin{enumerate}
%   \item Initialize $\boldsymbol{\Theta}$ randomly;
%   \item Initialize $\MATabund$ with $\boldsymbol{0}$.
% \end{enumerate}
    \algocomment{ADMM iterations} \nonumber
    \FOR{$k=1,\cdots,K$}
      %Choose a proper autoencoder network;
  %Initialize the weights of linear part of decoder with conventional endmember extraction methods such as VCA and parameters of other layers randomly;
 \algocomment{Updating $\bTheta$}
  \FOR{$i=1, \cdots,\textrm{epochs}$}
  \STATE Update $\mathsf{E}_{\bTheta_{\sf E}}(\bfW)$ via forward propagation,
  \STATE Compute the loss function by~\eqref{eq_loss},
  \STATE Update $\bTheta^{(k)}$ via retropropagation,
 \ENDFOR
 \algocomment{Updating $\MATabund$}
 \STATE Set $\MATabund^{(k-1,0)} = \MATabund^{(k-1)}$
 \FOR{$j=1, \cdots,J$ }
 %\STATE Apply the denoising operator $\mathsf{C}\left(\MATabund^{(k-1,j-1)}\right)$  
 \STATE Update $\MATabund^{(k-1,j)}$ with~\eqref{eq_mdoel_2D_RED_Ag_f},
 \ENDFOR
 \STATE Set $\MATabund^{(k)} = \MATabund^{(k-1,J)}$
 \algocomment{Updating $\mathbf{G}$}
 \STATE Update $\mathbf{G}^{(k)}$ with~\eqref{eq_mdoel_2D_RED_g};
    \ENDFOR
   %\ENDWHILE
   \ENSURE Estimated abundances $\MATabund$ and endmembers $\MATend$.
  \end{algorithmic}
\end{algorithm}

\section{Experimental results}
\label{sec:experiment}
This section presents experiments conducted to evaluate the effectiveness of the proposed unmixing framework. These experiments have been conducted on  synthetic and real data sets to quantitatively assess the unmixing results and  to demonstrate the effectiveness of our proposed method in real applications, respectively (see Sections \ref{subsec:expe_synthetic} and \ref{subsec:expe_real}). \\

\noindent \textbf{Compared methods -- } Several state-of-the-art methods have been compared. A first family of unmixing algorithms are conventional methods. SUnSAL-TV~\cite{iordache2012total} leverages on a handcrafted TV-term to regularize the optimization function. PnP-NMF~\cite{zhao2021hyperspectralNMF} is an NMF-based unmixing method, and denoisers are embedded as PnP to introduce prior information. A second family of compared methods is based on deep learning. CNNAE~\cite{palsson2020convolutional} is a deep AE-based unmixing method where convolutional filters capture spatial information. UnDIP~\cite{rasti2021undip} is a DIP-based unmixing method which uses a convolutional network. A geometric endmember extraction method is applied to estimate endmembers. SNMF~\cite{xiong2021snmf} is a deep unrolling algorithm, which unfolds the $\ell_p$-sparsity constrained NMF model into trainable deep architectures. CyCU-Net~\cite{gao2021cycu} proposes a cascaded AEs for unmixing with a cycle-consistency loss to enhance the unmixing performance.\\

\noindent \textbf{Hyperparameter settings -- }
All hyperparameters of the compared methods have been manually adjusted to obtain the best unmixing performance. The choice of the parameters associated with the proposed AE-RED method are discussed in detail as follows. The regularization parameters $\lambda$ and $\mu$ have been selected according to the noise level of generated data. More precisely,  $\lambda$ and  $\mu$ have been set to $0.5$ for the data with noise levels of $5$dB and $10$dB, to $0.1$ for the data with a noise level of $20$dB, to $0.01$ for the data with a noise level of $30$dB.
% Moreover, how these regularization parameters affect the unmixing results will be presented in the next subsection.
The learning rate to train the deep networks is set to $1\times10^{-3}$, and set $1\times10^{-4}$ to fine-tune the decoder weights. For the proposed CNN based unmixing method, the number $K$ of ADMM iterations is set to $15$, the number of epochs is set to $250$ and the number of inner iterations when updating the abundances is set to $J=1$. As for the proposed DIP based unmixing method, $K$, the number of epochs and $J$ are respectively set to $10$, $2300$ and $1$.\\
%denoiser
%parameter

\noindent \textbf{Performance metrics --} The root mean square error (RMSE) is used to evaluate the abundance estimation performance, which can be expressed by
  \begin{equation}\label{eq.rmse}
  \text{RMSE} = \sqrt{\frac{1}{NR}\sum_{i=1}^{N}\|\mathbf{a}_i-\hat{\mathbf{a}}_i\|^2},
\end{equation}
where $\mathbf{a}_i$ is the actual abundance of the $i$th pixel, and $\hat{\mathbf{a}}_i$ is the corresponding estimate. A smaller value of RMSE indicates better abundance estimation results. The endmember estimation is assessed by computing the mean spectral angle distance (mSAD) and the mean spectral information divergence (mSID) given by
  \begin{equation}\label{eq.sad}
  \text{mSAD}=\frac{1}{R}\sum_{r=1}^{R}\arccos\left(\frac{\Vend_{r}^{\top}\hat{\Vend}_r}
{\|\Vend_{r}\|\|\hat{\Vend}_r\|}\right)
\end{equation}
and
\begin{equation}\label{eq.sid}
  \text{mSID}=\frac{1}{R}\sum_{r=1}^{R}
\mathbf{p}_r\log\left(\frac{\mathbf{p}_r}{\hat{\mathbf{p}}_r}\right),
\end{equation}
where $\Vend_{r}$ and $\hat{\Vend}_r$ are the actual and estimate of the $r$th endmember, respectively, $\mathbf{p}_r={\Vend_r}/{\boldsymbol{1}^{\top}\Vend_r}$ and $\hat{\mathbf{p}}_i={\hat{\Vend}_r}/{\boldsymbol{1}^{\top}\hat{\Vend}_r}$. A smaller value indicates better estimation results. Finally, the peak signal-to-noise ratio (PSNR) is used to evaluate the image denoising and reconstruction, which is defined by
  \begin{equation}\label{eq.psnr}
  \text{PSNR}=10\times\log_{10}\left(\frac{\text{MAX}^2}{\text{MSE}}\right)
\end{equation}
where $\text{MAX}$ is the maximum pixel value of the reconstructed image $\hat{\mathbf{Y}}$ and $\text{MSE}$ is the mean square
error between the reconstructed image and the noise-free image. A higher value of PSNR indicates better reconstruction.

\begin{figure*}[!t]
  \centering
  \includegraphics[width=18cm]{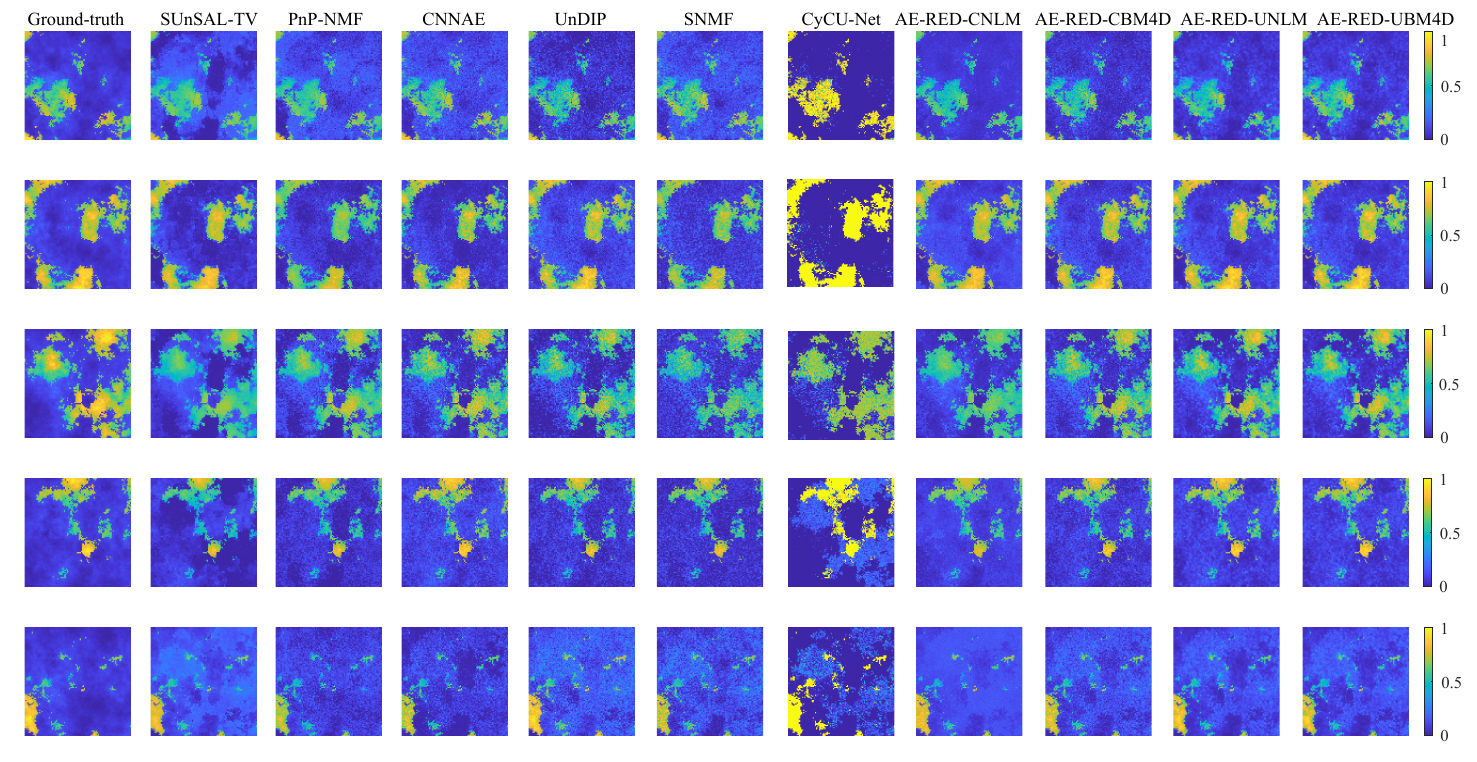}\\
  \caption{Synthetic data, SNR$=10$dB: estimated abundance maps.% (From Top to Bottom) Different endmembers. (From Left to Right) Ground-truth, SUnSAL-TV, PnP-NMF, CNNAE, UnDIP, SNMF, CyCU-Net, AEPnP-CNLM, AEPnP-CBM4D, AEPnP-UNLM and AEPnP-UBM4D.
  }\label{fig.abu_map_syn}
\end{figure*}

\begin{table}[!t]
%\scriptsize
\centering
\renewcommand\arraystretch{1.5}
\caption{synthetic data: performance (in term of RMSE) of the abundance estimation. Best RMSEs are reported in bold and underlined numbers denote the second best RMSEs.}\label{Tab_syn_RMSE_results}
\begin{tabular}{c|c|c|c|c}
\hline\hline
            & $5$dB       & $10$dB      & $20$dB      & $30$dB      \\ \hline
SUnSAL-TV   & 0.1112          & 0.0804          & 0.0284          & 0.0104          \\ \hline
PnP-NMF     & 0.1029          & 0.0711          & 0.0311          & 0.0117          \\ \hline
CNNAE       & 0.1078          & 0.0682          & 0.0292          & 0.0127          \\ \hline
UnDIP       & 0.1469          & 0.0854          & 0.0280          & 0.0100          \\ \hline
SNMF        & 0.1207          & 0.0906          & 0.0313          & 0.0112          \\ \hline
CyCU-Net    & 0.1150          & 0.0708          & 0.0296          & 0.0139          \\ \hline
AE-RED-CNLM  & 0.0943          & 0.0640          & 0.0261          & 0.0097          \\ \hline
AE-RED-CBM4D & 0.1009          & 0.0665          & \textbf{0.0235} & \textbf{0.0093} \\ \hline
AE-RED-UNLM  & \textbf{0.0919} & {\ul 0.0602}    & {\ul 0.0241}    & 0.0095          \\ \hline
AE-RED-UBM4D & {\ul 0.0972}    & \textbf{0.0585} & 0.0251          & {\ul 0.0094}    \\ \hline\hline
\end{tabular}
\end{table}

\begin{table}[h]
%\scriptsize
\centering
\renewcommand\arraystretch{1.5}
\caption{Synthetic data: performance (in term of mSAD) of the endmember estimation. Best mSADs are reported in bold and underlined numbers denote the second best mSADs.}\label{Tab_syn_mSAD_results}
\begin{tabular}{c|c|c|c|c}
\hline\hline
            & $5$dB            & $10$dB           & $20$dB           & $30$dB           \\ \hline
SUnSAL-TV   & 0.1013          & 0.0623          & 0.0173          & 0.0052          \\ \hline
PnP-NMF     & 0.0855          & 0.0533          & 0.0181          & 0.0083          \\ \hline
CNNAE       & 0.0811          & 0.0481          & 0.0162          & 0.0045          \\ \hline
UnDIP       & 0.0977          & 0.0685          & 0.0193          & 0.0057          \\ \hline
SNMF        & 0.0852          & 0.0595          & 0.0113          & 0.0043          \\ \hline
CyCU-Net    & 0.0826          & 0.0569          & 0.0146          & 0.0069          \\ \hline
AE-RED-CNLM  & 0.0769          & 0.0437          & \textbf{0.0103} & 0.0041          \\ \hline
AE-RED-CBM4D & 0.0770          & \textbf{0.0430} & {\ul 0.0105}    & 0.0042          \\ \hline
AE-RED-UNLM  & \textbf{0.0767} & 0.0434          & 0.0108          & {\ul 0.0040}    \\ \hline
AE-RED-UBM4D & {\ul 0.0768}    & {\ul 0.0433}    & 0.0107          & \textbf{0.0039} \\ \hline\hline
\end{tabular}
\end{table}

\begin{table}[h]
%\scriptsize
\centering
\renewcommand\arraystretch{1.5}
\caption{Synthetic data: performance (in term of mSID) of the endmember estimation. Best mSIDs are reported in bold and underlined numbers denote the second best mSIDs.}\label{Tab_syn_mSID_results}
\begin{tabular}{c|c|c|c|c}
\hline\hline
            & $5$dB            & $10$dB           & $20$dB           & $30$dB           \\ \hline
SUnSAL-TV   & 0.0391          & 0.0120          & 0.0013          & {\ul 0.0002}    \\ \hline
PnP-NMF     & 0.0195          & 0.0100          & 0.0011          & {\ul 0.0002}    \\ \hline
CNNAE       & 0.0432          & 0.0069          & 0.0013          & 0.0003          \\ \hline
UnDIP       & 0.0650          & 0.0130          & 0.0023          & \textbf{0.0001} \\ \hline
SNMF        & 0.1369          & 0.0112          & 0.0007          & \textbf{0.0001} \\ \hline
CyCU-Net    & 0.0447          & 0.0052          & 0.0014          & 0.0003          \\ \hline
AE-RED-CNLM  & \textbf{0.0184} & 0.0038          & \textbf{0.0005} & \textbf{0.0001} \\ \hline
AE-RED-CBM4D & 0.0189          & \textbf{0.0036} & {\ul 0.0006}    & \textbf{0.0001} \\ \hline
AE-RED-UNLM  & 0.0195          & {\ul 0.0037}    & 0.0007          & \textbf{0.0001} \\ \hline
AE-RED-UBM4D & {\ul 0.0187}    & {\ul 0.0037}    & {\ul 0.0006}    & \textbf{0.0001} \\ \hline\hline
\end{tabular}
\end{table}

\begin{table}[h]
%\scriptsize
\centering
\renewcommand\arraystretch{1.5}
\caption{Synthetic data: performance (in term of PSNR) of the image reconstruction. Best PSNRs are reported in bold and underlined numbers denote the second best PSNRs.}\label{Tab_syn_PSNR_results}
\begin{tabular}{c|c|c|c|c}
\hline\hline
            & $5$dB             & $10$dB            & $20$dB            & $30$dB            \\ \hline
SUnSAL-TV   & 30.8279          & 35.3531          & 43.9418          & 54.4288          \\ \hline
PnP-NMF     & 31.6765          & 36.2873          & 44.3496          & 54.6453          \\ \hline
CNNAE       & 31.4510          & 35.2500          & 43.3465          & 50.9367          \\ \hline
UnDIP       & 30.3016          & 34.8235          & 44.3141          & 54.6996          \\ \hline
SNMF        & 28.1358          & 32.2243          & 41.2482          & 51.3990          \\ \hline
CyCU-Net    & 30.8153          & 35.4829          & 42.6938          & 50.1539          \\ \hline
AE-RED-CNLM  & \textbf{32.4931} & {\ul 36.8916}    & 44.4119          & 54.7001          \\ \hline
AE-RED-CBM4D & 31.7141          & 36.1305          & \textbf{45.3226} & 54.8153          \\ \hline
AE-RED-UNLM  & 32.0177          & 36.6815          & 44.4916          & {\ul 55.0293}    \\ \hline
AE-RED-UBM4D & {\ul 32.2841}    & \textbf{36.9038} & {\ul 44.5306}    & \textbf{55.2367} \\ \hline\hline
\end{tabular}
\end{table}
\begin{figure}[h]
  \centering
  \includegraphics[width=8cm]{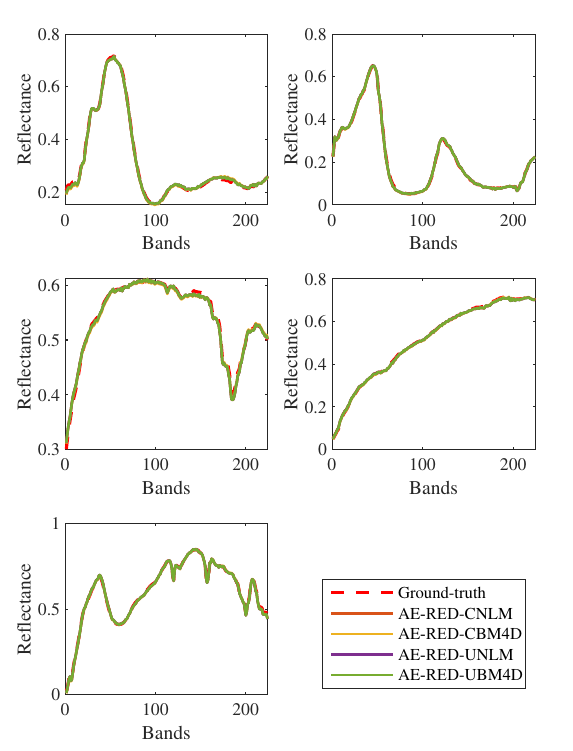}\\
  \caption{Estimated endmembers of synthetic data (SNR = 20 dB).}\label{fig.end_syn}
\end{figure}

\subsection{Experiments on the Synthetic data set}\label{subsec:expe_synthetic}
\noindent \textbf{Data description --} The synthetic images are composed of $100\times 100$ pixels. Abundance maps are generated using the method of the Hyperspectral Imagery Synthesis tools\footnote{http://www.ehu.es/ccwintco/index.php/Hyperspectral Imagery Synthesis
tools for MATLAB} to mimic a spatial homogeneity. A Gaussian field is drawn to generate the abundance matrix $\MATabund$. The abundance ground-truth is shown in Fig.~\ref{fig.abu_map_syn}.
The abundances maps satisfy ANC and ASC. Sets of $R=5$ endmembers are randomly selected from the U.S. Geological Survey (USGS) spectral library with a number of spectral bands of $B=224$. These endmembers are mixed according to the LMM and an additive zero-mean Gaussian noise is considered with variances according to $4$ signal-to-noise (SNR) ratios, i.e., $\mathrm{SNR} \in \left\{5\mathrm{dB}, 10\mathrm{dB}, 20\mathrm{dB}, 30\mathrm{dB}\right\}$.

\noindent \textbf{Results -- }
Tables~\ref{Tab_syn_RMSE_results}-\ref{Tab_syn_PSNR_results} report the estimation results obtained by the compared algorithms in terms of RMSE for the abundance estimation, mSAD and mSID for the endmember estimation and PSNR for the reconstruction. Conventional unmixing methods, such as SUnSAL-TV and PnP-NMF, achieve good unmixing results, demonstrating the usefulness of the explicit prior provided by manually designed regularization. Deep learning-based methods, such as CNNAE, SNMF and CyCU-Net, they can obtain suitable unmixing results and better endmember estimation results compared with the conventional methods, illustrating the ability of deep networks of embedding prior information. These results also show that the proposed AE-RED framework outperforms the compared state-of-the-art methods, across all performance metrics and the noise levels. Fig.~\ref{fig.abu_map_syn} depicts the estimated abundance maps associated with the synthetic data set with SNR$=10$dB. It can be observed that the abundance maps estimated by the AE-RED framework exhibit better agreement with the ground-truth. Fig.~\ref{fig.end_syn} shows the endmember estimated by the proposed framework on the synthetic data set with SNR$=20$dB, which are consistent with the ground-truth.

% convergence analysis, figure
% parameter analysis, figure

\subsection{Experiments on the Real data set}\label{subsec:expe_real}
\noindent \textbf{Data description --} Finally, experiments conducted on  two real data sets are discussed. Firstly, one considers the Samson data set, which was acquired by SAMSON observer and contains $B=156$ spectral channels ranging from $400$ nm to $889$ nm. The original image is of size of $952\times 952$ pixels, and a subimage of $95\times 95$ pixels is cropped in the experiment.  There are three endmembers in this data, namely ``water", ``tree" and ``soil". The second real data set used in these experiments is known as the Jasper Ridge image. It was acquired by Analytical Imaging and Geophysics (AIG) in 1999 with $B=224$ spectral bands covering a spectral range from $380$ nm to $2500$ nm. One considers a subimage of size of $100\times 100$ pixels and $B=198$ channels  after removing the bands affected by water vapor and atmospheric effects. It contains $R=4$ endmembers, namely ``water", ``soil", ``tree" and ``road".\\

\setlength{\tabcolsep}{3pt}
\begin{table*}[h]
\scriptsize
\centering
\renewcommand\arraystretch{1.5}
\caption{Samson and Jasper Ridge data sets: performance comparison in term of PSNR. Best results are reported in bold and underlined numbers denote the second best results.}\label{Tab_samson_results}
\begin{tabular}{c|c|c|c|c|c|c|c|c|c|c}
\hline\hline
 & SUnSAL-TV & PnP-NMF & CNNAE & UnDIP & SNMF & CyCU-Net & AE-RED-CNLM & AE-RED-CBM4D & AE-RED-UNLM & AE-RED-UBM4D \\ \hline
Samson & 32.6306 & 35.2650 & 28.6785 & \textbf{36.2918} & 29.7175 & 31.1702 & {\ul 35.6806} & 35.3954 & 35.5137 & 35.4009 \\ \hline
Jasper Ridge & {31.1325} & 29.2783 & 24.9619 & \textbf{32.1480} & 28.8746 & 28.7161 & 31.3895 & 30.0416 & {\ul 31.6140} & 30.0416 \\ \hline\hline
\end{tabular}
\end{table*}

\begin{figure*}[!t]
  \centering
  \includegraphics[width=18cm]{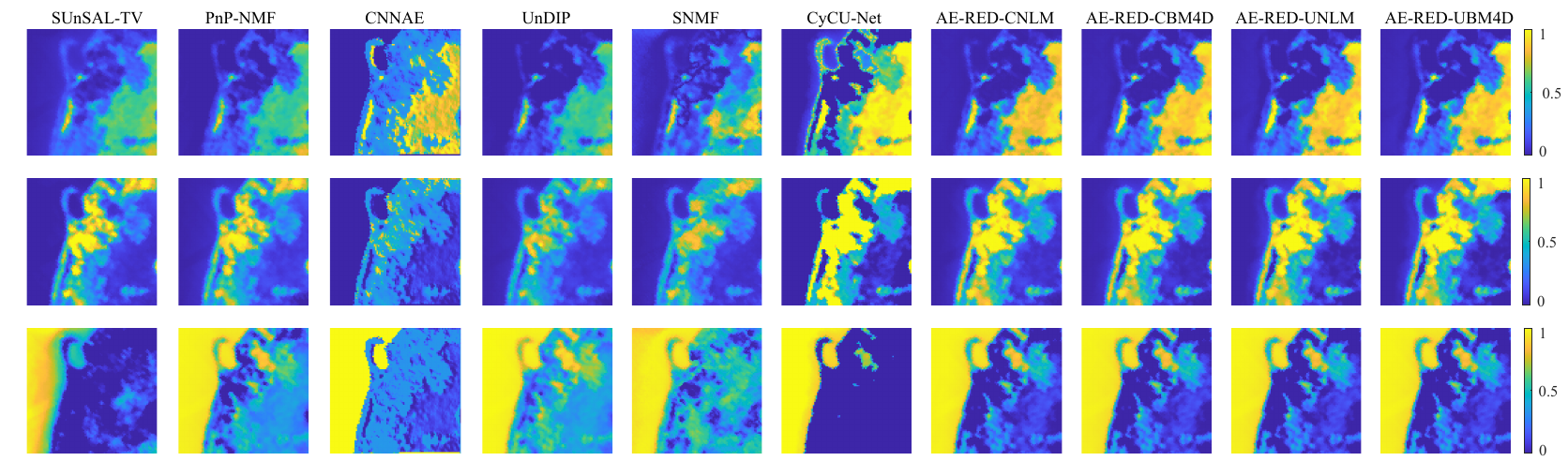}\\
  \caption{Samson data set: estimated abundance maps.% of . (From Top to Bottom) Different endmembers. (From Left to Right) SUnSAL-TV, PnP-NMF, CNNAE, UnDIP, SNMF, CyCU-Net, AEPnP-CNLM, AEPnP-CBM4D, AEPnP-UNLM and AEPnP-UBM4D.
  }\label{fig.abu_map_samson}
\end{figure*}

\begin{figure*}[!t]
  \centering
  \includegraphics[width=18cm]{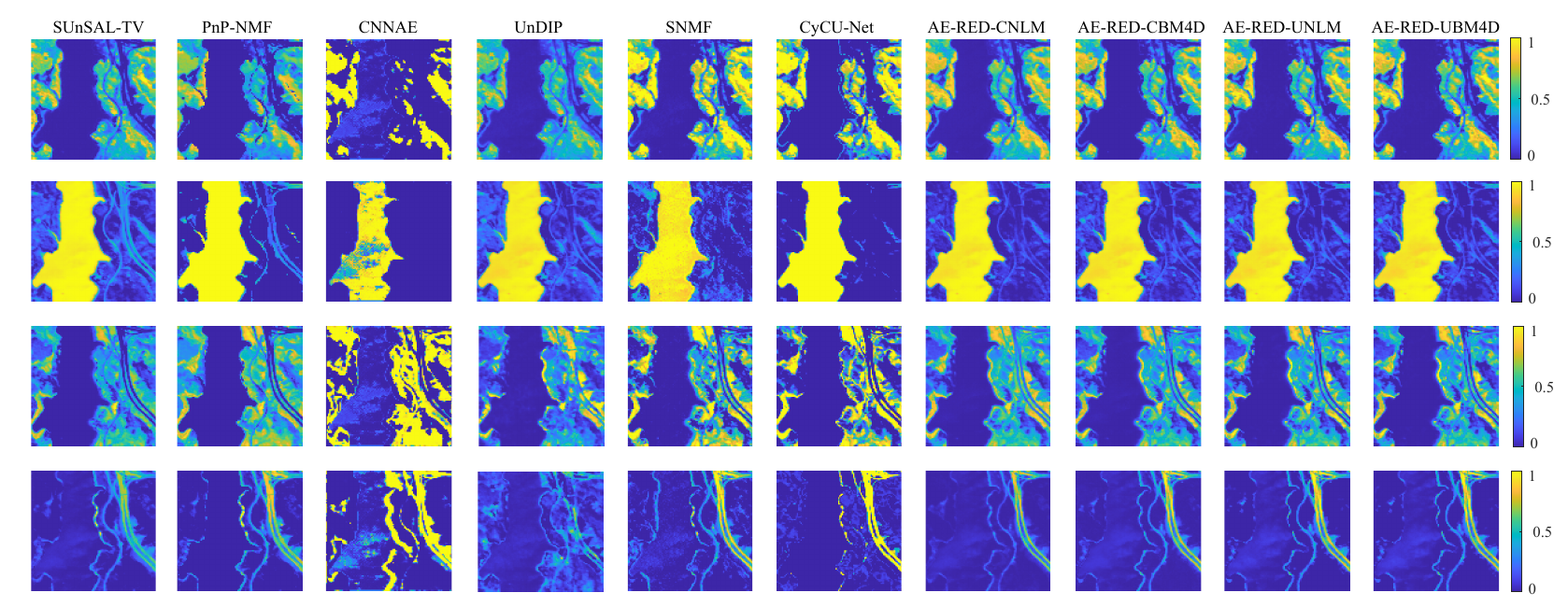}\\
  \caption{Jasper Ridge data set: estimated abundance maps% of . (From Top to Bottom) Different endmembers. (From Left to Right) SUnSAL-TV, PnP-NMF, CNNAE, UnDIP, SNMF, CyCU-Net, AEPnP-CNLM, AEPnP-CBM4D, AEPnP-UNLM and AEPnP-UBM4D.}
  \label{fig.abu_map_jas}
  }
\end{figure*}

\noindent \textbf{Results --} 
 As there is no available ground-truth for real datasets, a quantitative performance evaluation of abundances and endmembers cannot be provided. Therefore, we only rely on PSNR to evaluate the results of the compared methods. Table~\ref{Tab_samson_results} presents the PSNR performance associated with the compared methods obtained on the Samson data set. It is noteworthy that all methods produce comparable PSNR results, except for CNNAE, SNMF, and CyCU-Net, which provide significantly worse reconstruction. Although there is no ground-truth for the abundances, we can visually inspect the maps. For illustration purposes, we show the abundance maps estimated by the compared methods in Fig.~\ref{fig.abu_map_samson}. The proposed AE-RED framework can successfully separate the materials and provide sharp abundance estimates.
 %achieving the best mSAD results, particularly for soil and tree materials.
% As there is no available ground-truth for real datasets, we can not provide a quantitative performance evaluation of abundances and endmembers. Thus we only use PSNR to evaluate the results.
% The PSNR performance associated with the compared methods obtained on the Samson data set are reported in Table~\ref{Tab_samson_results}. 
% %It can be observed that the endmembers extracted by the proposed method achieve the best mSAD results, especially for the soil and tree materials. 
% It can be observed that all methods provide comparable PSNR, except CNNAE, SNMF and CyCU-Net which provides significantly worse reconstruction. 
% %Since there is no ground-truth for the abundances, only visual inspection of the maps is possible. 
% For illustrative purpose, the abundance maps estimated by the compared methods are shown in Fig.~\ref{fig.abu_map_samson}. The proposed AEPnP framework can successfully separate the materials and provide sharp abundance estimates.

Table~\ref{Tab_samson_results} also lists the PSNR results for the Jasper Ridge data set. 
%The proposed framework provides consistently the best SAD results for all materials. 
It can also be observed that the proposed method reaches the best PSNR. Fig.~\ref{fig.abu_map_jas} depicts the abundance maps estimated by all compared methods. Some of them, such as UnDIP, fail to recover the road. Due to the learning ability of deep networks, most deep learning based methods are able distinguish the individual materials. Finally the proposed AE-RED framework provides abundance maps with more detailed information and sharper boundaries.

\section{Conclusion}
\label{sec:conclusion}
This paper proposed a generic unmixing framework to embed a RED within an autoencoder. By carefully designing the encoder and the decoder, the autoencoder was able to provide estimated abundance maps and endmember spectra. In particular, for illustration purpose, two different encoder architectures are considered, namely a CNN and a DIP. Moreover the decoder could be chosen according to a particular mixture model. Leveraging on ADMM scheme, the resulting optimization problem was split into simpler subproblems. The first one was described by an objective function composed of a data-fitting term and a quadratic regularization. It was solved through the training of an autoencoder. The second subproblem was a standard RED objective function and solved by the fixed-point strategy. Two denoisers were considered, namely NLM and BM4D. The effectiveness of the proposed framework was evaluated through experiments conducted on synthetic and real data sets. The results showed that the proposed framework outperformed state-of-the-art methods. Future works include considering explicit endmember priors within the proposed framework and automatically selecting mixing model.

\bibliographystyle{IEEEtran}
\bibliography{IEEEfull,BIB}\ %IEEEabrv instead of IEEEfull

\end{document}